\pdfoutput= 1
\documentclass[11pt]{article}
\PassOptionsToPackage{sort&compress}{natbib}
\usepackage[final]{latex/acl}

\usepackage{times}
\usepackage{latexsym}

\usepackage[T1]{fontenc}
\usepackage[utf8]{inputenc}

\usepackage{microtype}

\usepackage{inconsolata}

\usepackage{graphicx}
\usepackage{amsmath} 
\title{Attention Consistency for LLMs Explanation}
\author{
\textbf{Tian Lan\textsuperscript{1,*}}, 
\textbf{Jinyuan Xu\textsuperscript{2,*}}, 
\textbf{Xue He\textsuperscript{3,4}}, 
\textbf{Jenq-Neng Hwang\textsuperscript{5}}, 
\textbf{Lei Li\textsuperscript{5,6,†}} \\
\\
\textsuperscript{1}Milkuya Studio \quad \textsuperscript{2}ERTIM, INALCO \quad \textsuperscript{3}Sorbonne University \\ \textsuperscript{4}IRD \quad \textsuperscript{5}University of Washington \quad \textsuperscript{6}VitaSight
\\
\\
}

\usepackage{float}
\usepackage{graphicx}
\usepackage{subcaption}

\usepackage{amssymb}
\usepackage{amsmath}
\usepackage{amsthm}
\usepackage{booktabs}
\usepackage{enumitem}
\usepackage{graphicx}
\usepackage{color}
\usepackage{microtype}
\usepackage{hyperref}
\usepackage{url}
\usepackage{booktabs}
\usepackage{graphicx}
\usepackage{lineno}
\usepackage{times}
\usepackage{hyperref}
\usepackage{algorithm}
\usepackage{algpseudocode}
\usepackage{multirow}
\usepackage{booktabs}
\usepackage{verbatim}
\usepackage{graphicx} % Make sure graphicx is loaded
\usepackage{caption}
\usepackage{booktabs}     % in your preamble
\usepackage{booktabs}
\usepackage{caption}
\usepackage{tabularx}
\usepackage{pifont}
\usepackage{xcolor}
\usepackage{graphicx}

\usepackage[dvipsnames]{xcolor} % Using dvipsnames for more color options like Lime
\usepackage[most]{tcolorbox}
\usepackage{soul} %
\newcommand{\emoji}[2][4.5ex]{%
  \raisebox{-0.2ex}{\includegraphics[height=#1]{#2}}%
}
\newcommand{\Good}{\textcolor{green}{\ding{51}}}   % ✓
\newcommand{\Bad }{\textcolor{red}{\ding{55}}}     % ✗
\definecolor{lightblue}{RGB}{173,216,230}
\definecolor{lightred}{RGB}{255,204,203}
\definecolor{lightgreen}{RGB}{144,238,144}
\sethlcolor{yellow!50} % Set a default highlight color for \hl
\tcbuselibrary{raster}
\algnewcommand{\Yield}{\State\textbf{yield}~} 

\begin{document}
\maketitle
\begin{abstract}

Understanding the decision-making processes of large language models (LLMs) is essential for their trustworthy development and deployment. However, current interpretability methods often face challenges such as low resolution and high computational cost. 
To address these limitations, we propose the \textbf{Multi-Layer Attention Consistency Score (MACS)}, a novel, lightweight, and easily deployable heuristic for estimating the importance of input tokens in decoder-based models. 
MACS measures contributions of input tokens based on the consistency of maximal attention. 
Empirical evaluations demonstrate that MACS achieves a favorable trade-off between interpretability quality and computational efficiency, showing faithfulness comparable to complex techniques with a 22\% decrease in VRAM usage and 30\% reduction in latency.

% MACS effectively balances efficiency and accuracy, with empirical evaluations showing performance comparable to complex techniques while incurring only a 10\% increase in VRAM usage and a 24\% reduction in inference speed.

% It offers an effective balance of efficiency and accuracy, empirical evaluations demonstrate that MACS achieves performance on par with complex techniques while introducing only a 10\% increase in VRAM usage and a 24\% reduction in inference speed compared to the baseline.

% This approach emphasizes sustained peak attention, producing clearer and more precise attribution maps.
%MACS measures input contributions by analyzing consistency in maximal attention, focusing on sustained peaks to generate clearer attribution maps.

%MACS offers an effective balance of efficiency and accuracy, introducing only a 10\% increase in VRAM usage and a 24\% reduction in inference speed compared to the baseline, while delivering competitive accuracy and faithfulness comparable to more resource-intensive methods. Empirical evaluations demonstrate that MACS achieves performance on par with complex techniques, making it an ideal tool for rapid inference-time diagnostics and a preliminary step for more intensive Explainable AI (XAI) analyses.

\end{abstract}

\section{Introduction}
The rapid advancements in Large Language Models (LLMs) have revolutionized natural language processing, powering diverse applications \citep{li2025human,he2025enhancing,jin2025visual,yao2025countllm,cai2025role,cai2025bayesian,li2025chatmotion}. As these models become increasingly integrated into critical systems, understanding their decision-making processes is crucial for ensuring trustworthiness, reliability, and enabling targeted improvements. Explainable AI (XAI), particularly methods attributing outputs to input tokens, is thus essential. However, existing approaches for interpreting deep decoder-only Transformers face notable limitations: full attention aggregation methods like Attention Rollout can produce noisy or diffuse attributions, potentially due to phenomena like Over-squashing, Over-mixing, and softmax dispersion which can obscure important signals in deep models and long sequences. Concurrently, many other XAI techniques, including gradient and perturbation-based approaches, are often too computationally intensive, limiting their practicality for real-time diagnostics and require model modifications or specialized computation paths.
Table \ref{tab:my_label} summarizes the comparison of MACS and these state-of-the-art XAI methods, outlining their respective advantages and limitations across key interpretability dimensions.  
\footnote{ * Both authors contributed equally to this research.}
\footnote{ † The corresponding author.}
\captionsetup[table]{font=small,skip=2pt}

% …
\begin{table}[t]
\setlength{\belowcaptionskip}{6pt}
\small
\centering
  \setlength{\tabcolsep}{2pt}        % column padding
  \renewcommand{\arraystretch}{1.1}  % slightly looser rows

  \begin{tabular}{@{}lcccc@{}}
    \toprule
    Metric             & \textbf{MACS}    & Rollout          & AttnLRP          & AtMan           \\
    \midrule
    Faithfulness       & \emoji{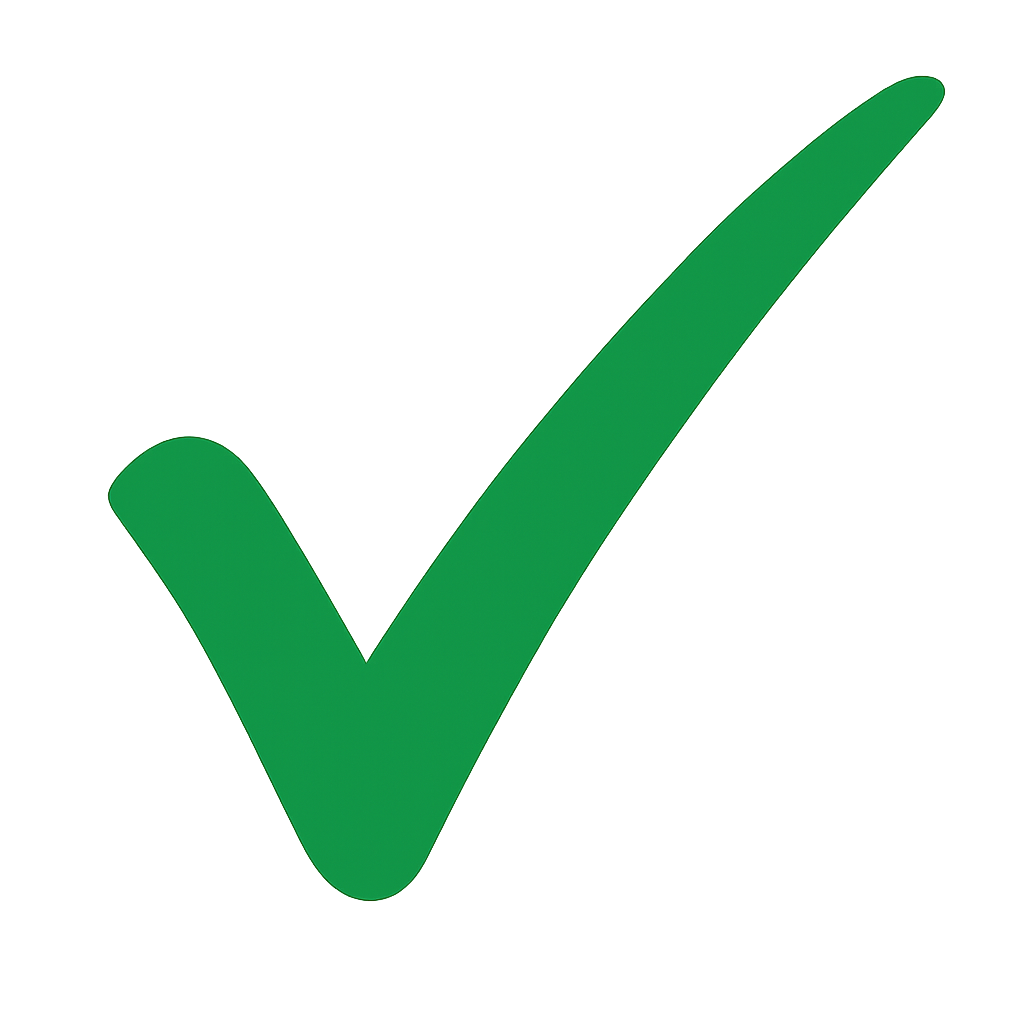}  & \emoji{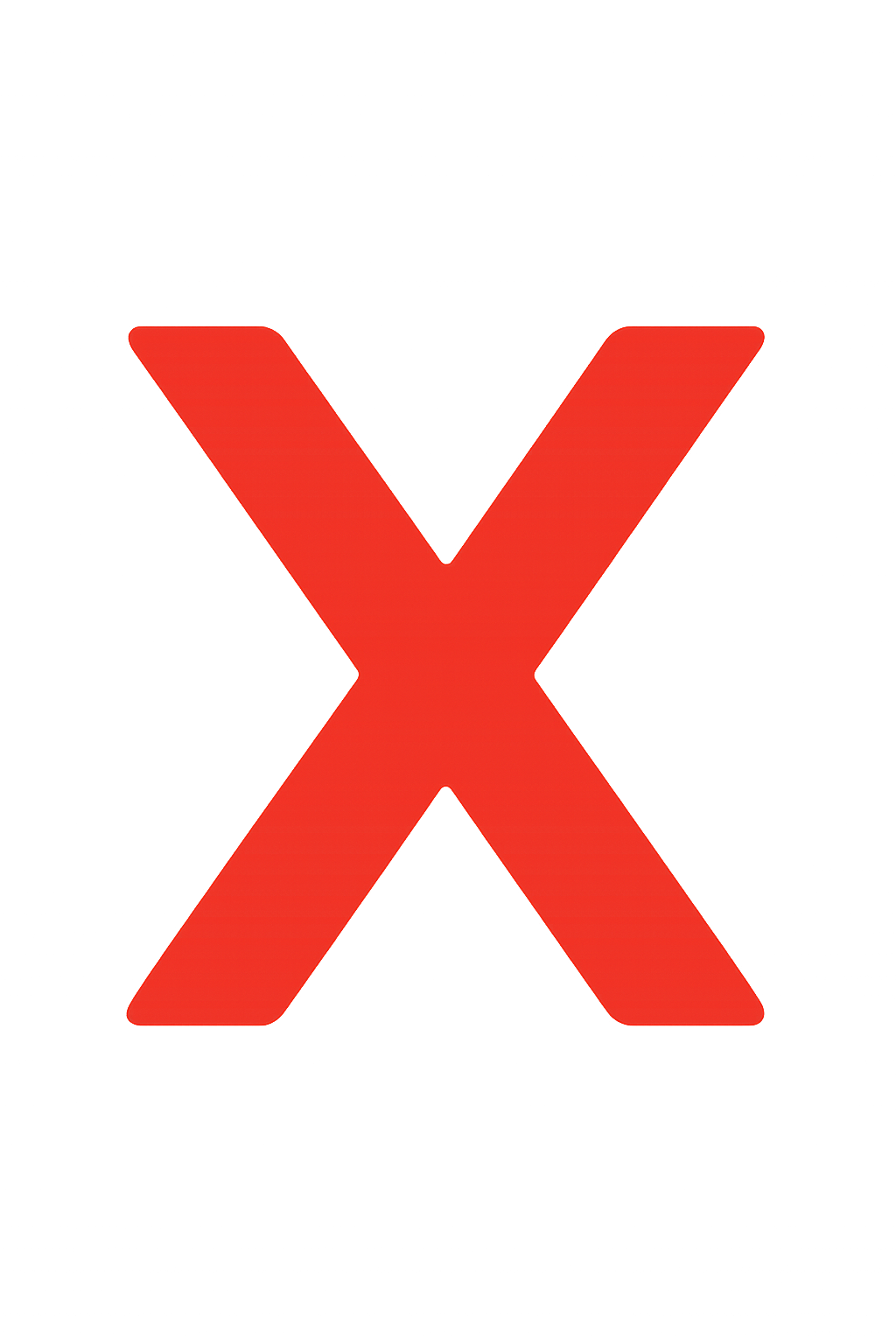}  & \emoji{Good.png} & \emoji{Bad.png}\\
    \addlinespace[3pt]
    Usability          & \emoji{Good.png} & \emoji{Good.png} & \emoji{Bad.png}  & \emoji{Bad.png} \\
    \addlinespace[3pt]
    Comp.~Eff.         & \emoji{Good.png} & \emoji{Bad.png} & \emoji{Good.png} & \emoji{Bad.png} \\
    \addlinespace[3pt]
    VRAM~Eff.          & \emoji{Good.png} & \emoji{Bad.png}  & \emoji{Good.png}  & \emoji{Good.png}\\
    \bottomrule
  \end{tabular}

  \caption{Comparative summary of MACS (ours) and other XAI methods, highlighting their strengths and weaknesses across key interpretability criteria. ({\Good}) denotes strong performance, ({\Bad}) indicates poor performance. MACS demonstrates competitive results across most metrics. }
  \label{tab:my_label}
\end{table}

To address the need for efficient and insightful interpretability, we propose the \textbf{Multi-Layer Attention Consistency Score (MACS)}, a novel, lightweight heuristic for estimating the importance of input tokens in Transformer models. Unlike conventional attribution techniques, MACS quantifies token contribution by evaluating the consistency of the strongest attention connection from an output query to each input token across all layers.  This consistency is derived via layer-wise max pooling followed by element-wise (Hadamard) multiplication. MACS can be applied directly during inference without requiring any model modifications.
% in your preamble:

Our main contributions are as follows:

\begin{itemize}
\item {\textbf{MACS Methodology}}: We propose the Multi-Layer Attention Consistency Score (MACS), a lightweight and computationally efficient heuristic for input token attribution in Transformer-based models, leveraging cross-layer attention consistency.

\item \textbf{Theoretical Motivation}: We provide a theoretical grounding for MACS as a response to known information propagation challenges (Over-squashing, Over-mixing, Softmax Dispersion), which often limit the effectiveness of existing attribution techniques.

\item \textbf{Empirical Evaluation}: Through experiments on Question Answering tasks, we demonstrate that MACS identifies salient tokens more effectively than Attention Rollout and achieves faithfulness comparable to more complex attribution methods.

\item \textbf{Practical Utility}: We demonstrate the real-time capability of MACS by highlighting its efficiency and showcase its potential applicability beyond text-only models to other Transformer architectures, as preliminarily explored in the context of Visual Question Answering (see Appendix \ref{sec:appendix b}).
\end{itemize}

We show that MACS achieves a favorable trade-off between computational efficiency and attribution quality, making it a practical component for interpreting LLMs.

\section{Background}
Understanding the decision-making process of Transformer-based models, especially Large Language Models (LLMs), is a critical research objective in the field of Explainable AI (XAI)~\cite{liu2025scene,shi2025explaining}. Existing attribution methods that link model predictions to input tokens can be broadly categorized by their primary mechanisms, including attention aggregation, perturbation, and backpropagation.
\subsection{Explainability via Attention Aggregation and its Challenges}

The attention mechanism has been a natural candidate for interpretability due to its explicit weighting of token interactions\cite{zhang2023attention}. Early studies demonstrated its usefulness in tasks such as machine translation \cite{zenkel2019addinginterpretableattentionneural} and document classification \cite{yang-etal-2016-hierarchical}. Subsequent analyses of models like BERT revealed that individual attention heads can capture linguistically meaningful patterns \cite{clark-etal-2019-bert}

However, interpreting raw attention from single layers or heads in deep, multi-head Transformers is often insufficient. To obtain a more holistic view, some methods focusing on attention aggregation across layers, like Attention Rollout \cite{abnar2020quantifyingattentionflowtransformers}, have been proposed. Rollout typically propagates attention by multiplying attention matrices across layers, thereby tracing and accumulating influence through all paths. Although more comprehensive than raw attention, such full-aggregation techniques may still yield "noisy" or diffuse attribution maps, making it challenging to identify critical input tokens.

Recent theoretical studies have highlighted challenges associated with aggregating attention signals in deep decoder-based Transformer models\cite{jia2024adaptive}. \citet{barbero2024transformersneedglassesinformation,barbero2025llmsattendtoken} formalized phenomena like Over-squashing, where influence from distant inputs is severely attenuated, and Over-mixing/Representational Collapse, where token representations lose distinctiveness. Furthermore, \citet{veličković2024softmaxforsharpoutofdistribution} demonstrate that standard softmax attention inherently disperses and loses its ability to focus sharply as sequence length increases out-of-distribution. These theoretical limitations suggest that methods like Rollout, by design, may struggle to produce clear attributions in deep, long-context models. This motivates the need for alternative heuristics like the one proposed in this work.

\subsection{Perturbation and Backpropagation-Based Explainability}
Beyond direct attention aggregation, prominent XAI paradigms include perturbation-based and gradient-based methods, each offering distinct approaches to attribution.

\textbf{Perturbation-based methods} assess feature importance by modifying input features or, in Transformer-specific variants like AtMan \cite{deiseroth2025atmanunderstandingtransformerpredictions}, by altering pre-softmax attention weights and measuring the impact on outputs. This family also includes local surrogate models such as LIME \cite{ribeiro2016whyitrustyou} and LORE \cite{guidotti2018localrulebasedexplanationsblack}, which approximate the model locally using interpretable proxies, and SHAP \cite{lundberg2017unifiedapproachinterpretingmodel}, which leverages game-theoretic principles. While generally model-agnostic, these methods require numerous forward passes, making them computationally expensive. Moreover, surrogate-based explanations may fail to capture the internal representations of the original model.

\textbf{Gradient-based methods} propagate signals from the output back to the input using gradient information. Techniques such as SmoothGrad \cite{smilkov2017smoothgradremovingnoiseadding} and Integrated Gradients \cite{sundararajan2017axiomaticattributiondeepnetworks} belong to this family, however their effectiveness can degrade in large Transformers \cite{achtibat2024attnlrpattentionawarelayerwiserelevance}. Other approaches, like gradient-weighted attention \cite{chefer2021transformerinterpretabilityattentionvisualization} (e.g., Grad-CAM \cite{Selvaraju_2019} variants), and rule-based techniques like Layer-wise Relevance Propagation (LRP) \cite{binder2016layerwiserelevancepropagationneural} and AttnLRP \cite{achtibat2024attnlrpattentionawarelayerwiserelevance}, provide more structured attributions but face challenges in handling Transformer-specific non-linearities while maintaining theoretical guarantees like relevance conservation.

Both families are typically post hoc and computationally demanding, often requiring backward passes or repeated evaluations. This limits their scalability for real-time or interactive scenarios, motivating the development of efficient inference-time alternatives like MACS.

\section{Methodology: From Full Aggregation to Attention Consistency}
This section outlines the motivation and formulation of our proposed method, MACS. We begin by discussing the limitations inherent in full attention aggregation approaches, which motivates our alternative strategy based on measuring attention consistency.
\subsection{More is Less: Limitations of Full Aggregation Methods}
Aggregation-based methods like Attention Rollout aim to approximate how influence propagates through a Transformer model. These methods operate on the principle that each layer uses self-attention to redistribute influence among token representations. From a graph-theoretic perspective,  the model can be viewed as a directed graph, where nodes correspond to token representations and edges represent attention-based interactions. Attention Rollout seeks to estimate the total influence flowing from an initial input token $j$ to a token representation $i$ at a later layer $L$ by aggregating attention patterns across layers.

A common formulation involves recursively multiplying attention matrices $A^{(l)}$ (often adjusted to account for residual connections):
\begin{equation}\label{eq:tildeA}
\tilde{A}^{(L)} = A^{(L)} A^{(L-1)} \cdots A^{(0)}
\end{equation}

The value $(\tilde{A}^{(L)})_{ij}$ provides an estimate of the total influence accumulated along all paths from input $j$ to position $i$ at layer $L$.

However, a criticism of such full aggregation methods is that the resulting attribution map can be "noisy" \cite{achtibat2024attnlrpattentionawarelayerwiserelevance}. These methods may highlight a large number of tokens, making it difficult to identify which ones were truly influential. Recent studies on the dynamics of deep decoder-only Transformers have identified several contributing factors:

\textbf{Over-squashing} \cite{barbero2024transformersneedglassesinformation,barbero2025llmsattendtoken}: In deep causal models, information propagating over long distances (many layers or across many tokens) can become progressively weaker or "squashed". Aggregating contributions from all paths, as Rollout does, means accumulating potentially numerous weak, squashed signals.

\textbf{Over-mixing / Representational Collapse} \cite{barbero2024transformersneedglassesinformation,barbero2025llmsattendtoken}: Through successive layers of transformation, token representations can become less distinct and more similar to each other. This may make attention patterns less discriminative, and aggregating them might lead to a blurred or averaged-out view of influence.

\textbf{Softmax Dispersion} \cite{veličković2024softmaxforsharpoutofdistribution}: The softmax attention mechanism inherently struggles to maintain sharp focus as the number of attended items grows, with attention weights dispersing towards uniform. This affects methods relying on aggregating these weights.

Because Attention Rollout structurally sums influence over all possible paths via matrix multiplication, it is susceptible to accumulating noise from these weak (squashed) or potentially indistinct (mixed) signals. This can result in dense or diffuse attribution maps where identifying the most critical input tokens for a specific output is challenging. This motivates our exploration of a more focused approach.

\subsection{Less is More: MACS - Measuring Attention Consistency Across Layers}
To provide an alternative perspective that might yield clearer attribution while remaining computationally efficient for real-time diagnostic purposes, we propose  \textbf{Multi-Layer Attention Consistency Score (MACS)}. 

Instead of summing contributions over all forward paths like Rollout, MACS adopts a different heuristic that focuses on measuring the \textbf{consistency of attention links} directed backward from the generated token's query to each input token's key, across all layers. 

The core idea is that input tokens maintaining a sustained, strong attention connection from the output query across the network's depth are likely key contributors. This provides a distinct measure of contribution based on the consistency of focused attention rather than total aggregated flow.  We hypothesize that emphasizing consistent, strong links can yield clearer and potentially sparser attribution maps by filtering out weak or inconsistent signals.

\subsubsection{Setup and Generation Process}
Given an initial input sequence $X = \{x_1, x_2, \dots, x_N\}$, where $x_i$ represents the embedding of the $i$-th input token, a decoder-only Transformer generates subsequent tokens autoregressively. The token $t_k$ (at sequence position $n = N+k$) is generated based on the conditional probability:
\begin{equation}\label{eq:tildeB}
\small
t_k \;\sim\; P\bigl(t_k \mid x_1, \dots, x_N,\; t_1, \dots,t_{k-2},t_{k-1}\bigr)
\end{equation}
Modern decoders utilize a caching mechanism for efficiency. At generation step $k$ (producing token $t_k$ at position $n=N+k$), only the representation of the most recently generated token $t_{k-1}$ (at position $n-1 = N+k-1$)\footnote{At the first generation step ($k=1$, where $n = N+1$), MACS is bootstrapped using the attention vector from the input’s last token $x_N$, representing the model’s attention pattern just before generation begins.} typically issues new queries into the attention mechanism. These queries attend to the keys and values associated with all preceding tokens in the sequence (both original inputs and previously generated tokens).

For MACS, we focus on the attention patterns generated when predicting token $t_k$ (at position $n$). Let $L$ be the number of layers in the Transformer (indexed $l=0$ to $L$). For each layer $l$ and head $h$, let $a_{uv}^{(l,h)}$ denote the attention weight from the query at position $u$ to the key at position $v$. The attention vector \(\mathbf{a}^{(n-1,l,h)} \;\in\;\mathbb{R}^{\,n-1}\) computed by the query from position $n-1$ (associated with predicting $t_k$) is:

\begin{multline}
\mathbf{a}^{(n-1,l,h)} = \mathrm{Attn}^{(l,h)}\Biggl( \underbrace{x_1,\dots,x_N}_{\text{keys: initial input}}, \\
\underbrace{t_1,\dots,t_{k-2},t_{k-1}}_{\text{keys: generated so far}} \Biggm| \text{query: } n-1 \Biggr)
\end{multline}

where $\mathrm{Attn}^{(l,h)}(\cdot)$ denotes the standard scaled dot-product attention computation for head $h$ at layer $l$ under causal masking.
\subsubsection{MACS Calculation Steps}
MACS processes these attention vectors layer-by-layer for each generation step $k$ (predicting token $t_k$ at position $n$) to compute a consistency score $C_i^{(k)}$ \footnote{For simplicity, we omit the step index $(k)$ in the layer-wise calculations and denote the token $i$ consistency score as $C_i$. The index $(k)$ will be reintroduced when presenting scores across multiple steps. } for each input token $i \in \{1, \dots, N\}$.

\textbf{Step 1: Attention Extraction and Redistribution}
\label{step1}
For each layer $l$ and head $h$:
\begin{itemize}
    \item Extract attention to inputs: 
    
    $\mathbf{a}_{I}^{(n-1,l,h)} \in \mathbb{R}^{N}$, where $(\mathbf{a}_{I}^{(n-1,l,h)})_i = a_{n-1,i}^{(l,h)}$ 
    for $1 \le i \le N$.
    \item Extract attention to previous outputs: 
    
    $\mathbf{a}_{O}^{(n-1,l,h)} \in \mathbb{R}^{k-1}$, where $(\mathbf{a}_{O}^{(n-1,l,h)})_p = a_{n-1, (N+p)}^{(l,h)}$ for $1 \le p \le k-1$.
    \item Calculate redistributed attention $\mathbf{a}_{R}^{(n-1,l,h)} \in \mathbb{R}^{N}$:
\begin{equation}
\small
(\mathbf{a}_{R}^{(n-1,l,h)})_i = a_{I, i}^{(n-1,l,h)} + \mkern-4mu \underbrace{\frac{1}{N} \sum_{p=1}^{k-1} a_{O, p}^{(n-1,l,h)} }_{\text{Avg. Attention to Outputs}} 
\end{equation} 

\end{itemize}

\textbf{Justification}: This step incorporates indirect influence ($t_{k-1} \to t_p \to i$) by uniformly distributing attention from previous outputs back to inputs, aiming to balance attention in long generations where direct attention to distant inputs might decay due to effects like over-squashing.

\textbf{Step 2: Max-Pooling Across Heads}
\label{step2}
For each layer $l$, compute the element-wise maximum across heads:
\begin{equation}
(\mathbf{m}'_{l})_i = \max_{h \in \{1, \dots, H\}} \left( (\mathbf{a}_{R}^{(n,l,h)})_i \right) \quad %\text{for } i = 1, \dots, N
\end{equation}
This yields $\mathbf{m}'_{l} \in \mathbb{R}^N$.

\textbf{Justification}: This step isolates the strongest attention signal directed toward input token $i$ across all heads at layer $l$. By focusing on peak signals from potentially specialized heads, it filters out weaker, diffuse attention often encountered in over-mixing scenarios, thereby highlighting the most decisive attention link per layer and mitigating noise from less informative heads.

\textbf{Step 3: Incorporate floor vector}
\label{step3}
Introduce a minimum score offset using hyperparameter $\alpha$ (e.g., 0.8):
\begin{equation}
    \mathbf{m}_{l} = \alpha \mathbf{m}'_{l} + (1 - \alpha) \mathbf{1}^N
\end{equation}
\textbf{Justification}: This step ensures that even if $\mathbf{m}'_{l}$ has near-zero entries (i.e., no head strongly attended to token $i$ at layer $l$), the score going forward $(\mathbf{m}_{l}$ is at least $1-\alpha$. This prevents the subsequent Hadamard product from prematurely zeroing out the contribution score for tokens whose relevance might only emerge in deeper layers.

\textbf{Step 4: Multi-layer Attention Consistency Score (MACS)}

Measure sustained layer-wise attention strength through layer-wise multiplication:

Initialize $\mathbf{c}_{0} = \mathbf{m}_{0}$. 

For $l = 1, \dots, L$:
\begin{equation}
\mathbf{c}_{l} = \mathbf{m}_{l} \odot \mathbf{c}_{l-1}
\end{equation}

The final consistency vector is $\mathbf{c}_{L} \in \mathbb{R}^N$ where $C_i^{} = (\mathbf{c}_{L})_i$

\textbf{Justification}: The Hadamard product directly measures the \textbf{consistency} of the processed maximal attention link across layers. For $(\mathbf{c}_{L})_i$ to be large, the corresponding $(\mathbf{m}_l)_i$ must be consistently large across most/all layers $l$. This contrasts with additive aggregation and aims to yield clearer attribution by emphasizing sustained relevance, filtering transient signals that could contribute to noise exacerbated by over-squashing or softmax dispersion.

\subsubsection{Final Attribution Scores via Z-Scoring}

The raw MACS score vector \(\mathbf{c}_{L}^{(k)} \in \mathbb{R}^N\) at generation step \(k\) captures each token’s layer-consistent maximal attention strength. To render these values comparable and highlight the most salient tokens at each step, we normalize them directly into Z-scores:
\begin{equation}
    z_i^{(k)} \;=\; \frac{\bigl(\mathbf{c}_{L}^{(k)}\bigr)_i \;-\; \mu^{(k)}}{\sigma^{(k)}},
\end{equation}
where \(\mu^{(k)}\) and \(\sigma^{(k)}\) are the mean and standard deviation across the $N$ input token scores within the vector \(\mathbf{c}_{L}^{(k)}\). Each \(z_i^{(k)}\) directly measures how many standard deviations token \(i\)’s attention consistency deviates from the average, immediately flagging tokens with statistically significant focus during generation.

\subsection{Algorithm Overview}

The complete MACS algorithm is summarized as follows:
\begin{algorithm}[H]
\caption{MACS (streaming)}
\label{alg:token_importance}
\begin{algorithmic}[1]
\For{each generation step $k$}
    \State Extract and redistribute attention
    \For{each transformer layer $l$}
        \State Max-pooling across attention heads
        \State Add weighted floor vector
        \State Calculate attention consistency
    \EndFor
    \State Compute step Z-score $z^{(k)}$
    \Yield $z^{(k)}$ 
    \EndFor
\end{algorithmic}
\end{algorithm}

\section{Experiments}
In the Experiments section, we address two key research questions. \textbf{First}, how does our proposed method compare to alternative approaches in terms of accuracy and reliability? \textbf{Second}, what is the impact of our method on the model's inference performance?

We perform all experiments on an NVIDIA A800 (80 GB) GPU, using \textbf{Llama 3.1-8B}  \cite{grattafiori2024llama3herdmodels} as our primary model. This choice was made because Llama 3.1-8B is a powerful, widely-used, and publicly available LLM whose architecture is characteristic of many current state-of-the-art decoder-only Transformers, ensuring our findings are broadly relevant to a significant class of contemporary LLMs. 

We benchmark MACS against four baseline methods representing distinct explainability paradigms:

\begin{itemize}
    \item \textbf{Random}: Assigns random importance scores, serving as a basic sanity check.
    \item \textbf{Attention Rollout}: An attention aggregation method that sums influence over all paths. Chosen as a widely recognized aggregation-based baseline.
    \item \textbf{AttnLRP}: A state-of-the-art gradient-based method adapted from Layer-wise Relevance Propagation for Transformers. Chosen as a strong, more complex gradient-based baseline.
    \item \textbf{AtMan}: A perturbation-based method that manipulates pre-softmax attention. Chosen to represent post hoc perturbation techniques.
\end{itemize}

The key hyperparameter $\alpha$ for MACS is set to 0.8.  For AttnLRP, we use their official \textit{lxt } \footnote{https://github.com/rachtibat/LRP-eXplains-Transformer} library, and for AtMan, we adapt their publicly available code\footnote{https://github.com/Aleph-Alpha-Research/AtMan} to Llama 3.1-8B.
%for the QA task and \textbf{Qwen 2.5-VL-7B Instruct}\cite{bai2025qwen25vltechnicalreport} for Visual QA

\subsection{Question Answering (QA) Task Attribution}
\subsubsection{Dataset, Metrics and Implementation Details}
We evaluate attribution performance on a Question Answering task using a subset of 350 question-context-answer triples from the SQuAD 2.0 dataset \cite{rajpurkar2018knowdontknowunanswerable}, where answers are guaranteed to be spans within the context.
we use the following prompt:
\begin{quote}
\texttt{Answer the question based on the following text. Keep your response short and simple. Do not quote the original text.\\
Question: \{question\}\\
Context: \{context\}}
\end{quote} 
This subset has an average context length of 169 tokens (max 512), and we set the maximum generation length to 256 tokens. If multiple answers exist, metrics are averaged.

Our primary evaluation metric for ranking relevant input tokens is the Area Under the Precision-Recall Curve (AUC-PR), which evaluates how well each method ranks the ground-truth answer tokens (answer spans) within the input context. The final AUC-PR reported for a given sample is the\textbf{ maximum AUC-PR achieved across all generation steps} for that sample. This "best step" approach acknowledges that an attribution method might highlight the answer most clearly at a particular point during the generation of the response.

To assess faithfulness, we adopt the Symmetric Relevance Gain (SRG) \cite{blücher2024decouplingpixelflippingocclusion}, which quantifies the difference between the model's performance (e.g., output text similarity to the original or model confidence) when progressively removing the least influential tokens versus removing the most influential tokens. We report SRG based on model perplexity (SRG-PP) and ROUGE-L F1 score (SRG-RL). For SRG-PP, more negative values indicate better faithfulness (as perplexity should ideally increase more with important token removal); for other metrics, higher is better. The SRG for the Random baseline is computed using two independent random perturbation orderings and is expected to be near zero. Confidence intervals (95\%) are reported as mean ± half-width. Full details and additional results are in Appendix \ref{sec:appendix a} and Table \ref{tab:comprehensive_results}.

\subsubsection{Results and Discussions}
\begin{table}[h]
\small
\setlength{\textfloatsep}{2pt}
\centering
\scriptsize                    % ← 这里把字号降到 
\setlength{\tabcolsep}{3pt}     % 列间距再紧一点
\renewcommand{\arraystretch}{1.0}
\footnotesize    % tighter horizontal padding
\renewcommand{\arraystretch}{1.1} % slightly reduced row height
\begin{tabular}{@{}lccc@{}}
\toprule
\textbf{Method} & \textbf{mAUC-PR ↑} & \textbf{mSRG-PP ↓} & \textbf{mSRG-RL ↑} \\
\midrule
Random        & 0.113 \tiny{± 0.01} & 0.003 \tiny{± 0.01} & -0.006 \tiny{± 0.01} \\
Rollout       & 0.147 \tiny{± 0.02} & -0.039 \tiny{± 0.02} & 0.082 \tiny{± 0.02} \\
AttnLRP       & 0.565 \tiny{± 0.03} & \textbf{-0.126} \tiny{± 0.01} & \textbf{0.323} \tiny{± 0.02} \\
AtMan         & 0.315 \tiny{± 0.03} & -0.021 \tiny{± 0.01} & 0.055 \tiny{± 0.02} \\
MACS(Ours) & \textbf{0.601} \tiny{± 0.03} & -0.118 \tiny{± 0.01} & 0.315 \tiny{± 0.02} \\
\bottomrule
\end{tabular}
\caption{Performance on the QA task. SRG = Symmetric Relevance Gain; PP = Perplexity; RL = ROUGE-L; higher/negative is better; The SRG for the Random baseline is computed using two independent random perturbation orderings and is expected to be near zero. ± value represents the half-width of the 95\% CI. m represents the averaged value across all samples.}

\label{tab:qa_attribution_results}
\end{table}

Table \ref{tab:qa_attribution_results} presents the attribution performance among the tested XAI methods. MACS leads in identifying relevant answer tokens, achieving an mAUC-PR of 0.601.This significantly outperforms Attention Rollout by over 300\% ($0.147 \pm 0.02$) and AtMan by over 90\% ($0.315 \pm 0.03$).

In terms of faithfulness, MACS demonstrates performance statistically comparable to the more complex AttnLRP method across both mSRG-RL and mSRG-PP. Both substantially outperform Rollout and AtMan. 

The poor performance of Attention Rollout across both mAUC-PR and faithfulness metrics suggests its tendency to produce diffuse or misleading attributions, a behavior consistent with the theoretical challenges of signal degradation from over-squashing and softmax dispersion in deep models. 

\subsubsection{Analysis of "Best Step" Timing}
\label{best_step}
To further understand the nature of the attention consistency captured by MACS, we analyzed when its "best step" (the generation step with highest AUC-PR for identifying the answer span in the context) occurs relative to the actual generation of the answer. Across our 350 SQuAD samples, we found that in \textbf{97.68\% of cases, MACS achieved its peak AUC-PR towards the answer span before the model began to generate the answer tokens themselves}. An illustrative example is shown in Figure \ref{fig:gshowcase}, where MACS highlights the answer in the context based on the query from an early generated token, prior to the answer appearing in the output. This finding suggests that MACS captures anticipatory attention related to information retrieval and comprehension crucial for answer formation, rather than merely reflecting a surface-level similarity during the token-by-token generation of the answer.

\subsection{Performance Comparison}
Beyond attribution quality, the practical utility of XAI methods hinges on their computational efficiency, especially for real-time applications or analysis of large models and long contexts. In this section, we evaluate the efficiency of MACS against other attribution methods.

\subsubsection{Dataset, Metrics, and Implementation Details}

We focus on two key aspects: peak VRAM usage and throughput (tokens generated per second). 
It’s important to distinguish how these metrics apply across XAI methods. For real-time capable methods such as MACS, AttnLRP, and Rollout, VRAM and throughput reflect overhead added during the generation process. In contrast, post-hoc methods like AtMan report "Peak VRAM" as the sum of baseline inference memory and the additional memory used for attribution, such as perturbations. Similarly, AtMan's "Throughput" is calculated using the total time, including both inference and perturbation, divided by the number of tokens generated in the original inference, capturing the full explanation cost.

To evaluate computational efficiency under demanding conditions, we sampled 70 CNN/DailyMail \cite{nips15_hermann} examples with input lengths ranging from 173 to 3,936 tokens, paired with a summarization task requiring up to 512 output tokens. This setup imposes substantial and diverse workloads, allowing for a clear assessment of VRAM usage and throughput scaling across different attribution methods as context length increases. The prompt used is as follows:

\begin{quote}
    \texttt{Summarize the following text.\\
    Text: \{context\}}
\end{quote}
\subsubsection{Results and Discussions}

\begin{figure*}[t]      % * 号表示跨双栏
  \centering
  \begin{subfigure}[b]{0.45\textwidth}   % 注意改为 textwidth
    \includegraphics[width=\linewidth]{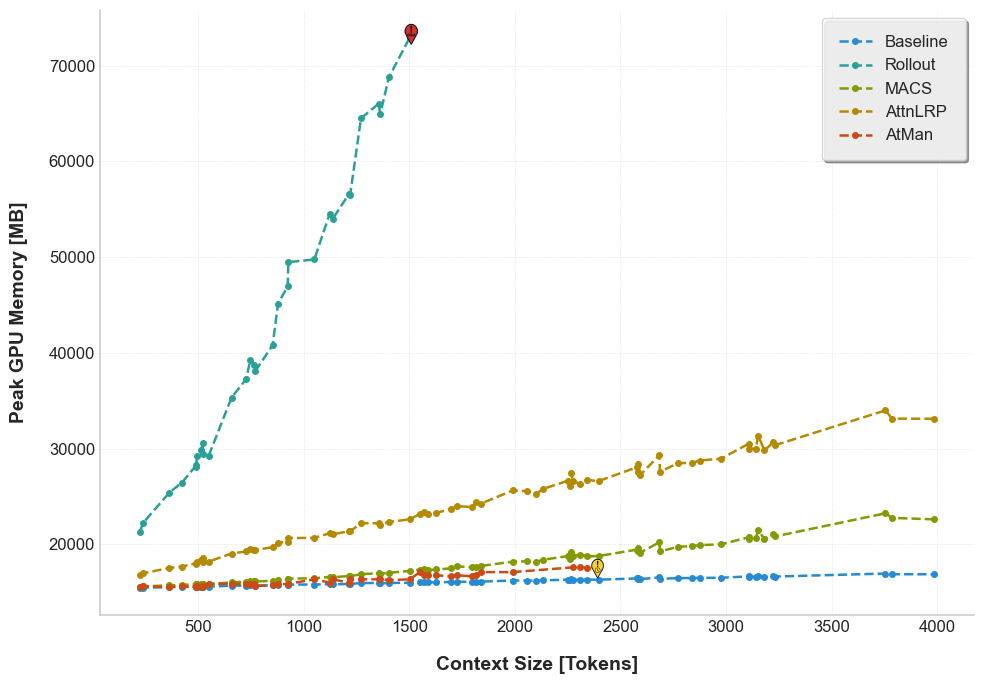}
    \subcaption{Peak VRAM vs Context Size}
    \label{fig:subfigA}
  \end{subfigure}
  \hfill
  \begin{subfigure}[b]{0.45\textwidth}
    \includegraphics[width=\linewidth]{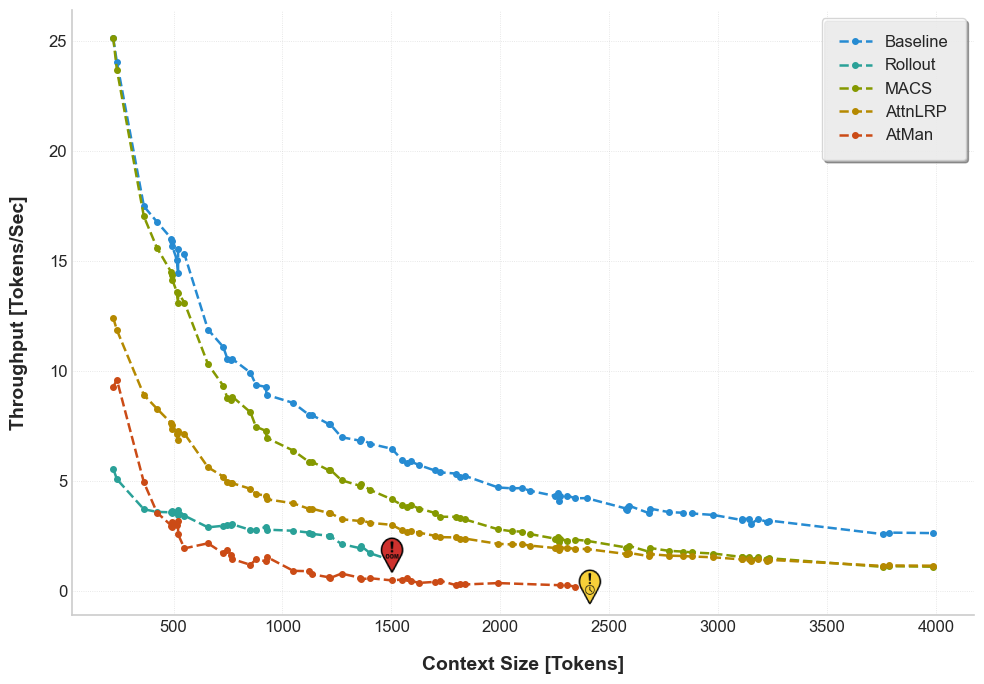}
    \subcaption{Throughput vs Context Size}
    \label{fig:subfigB}
  \end{subfigure}
  \caption{Peak VRAM usage (a) and Throughput (b) against context size for different XAI methods. 
    \emoji[2ex]{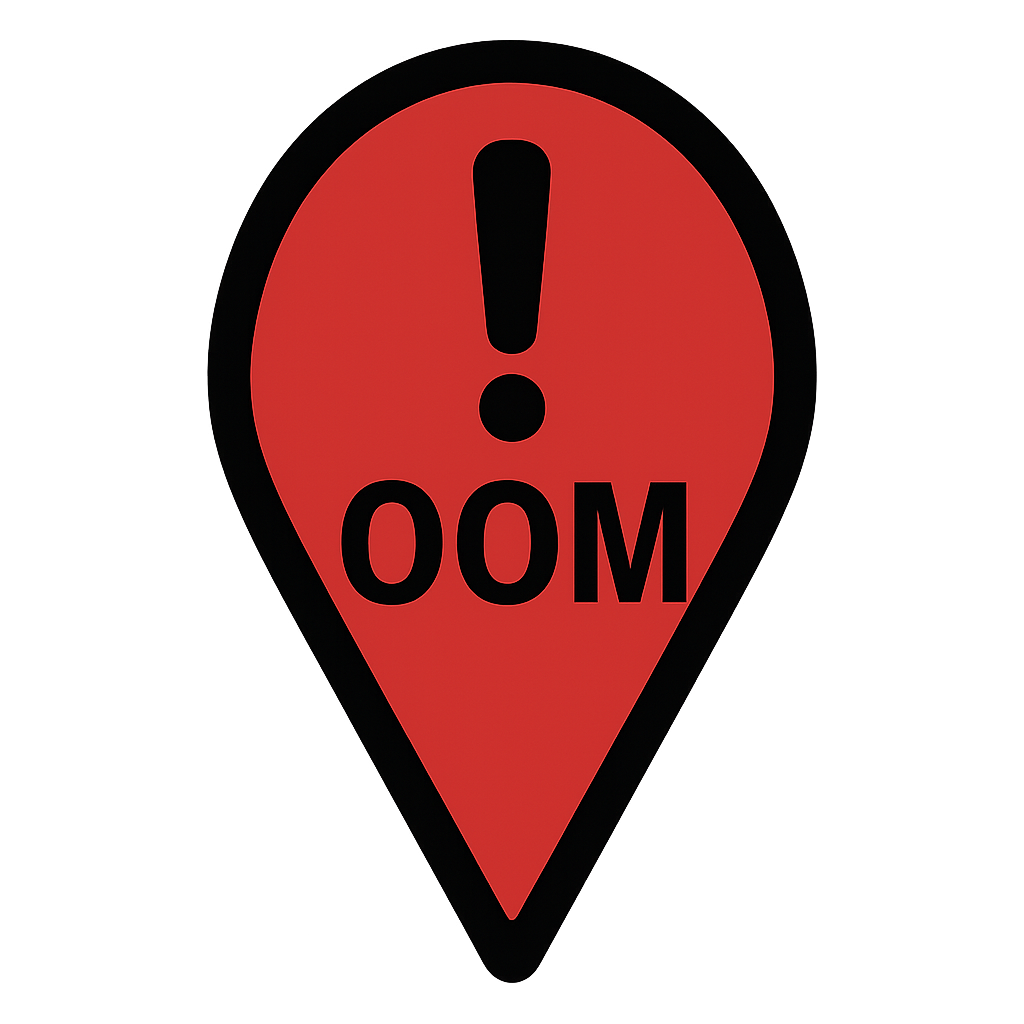} denotes Out-of-Memory (OOM) errors. 
    \emoji[2ex]{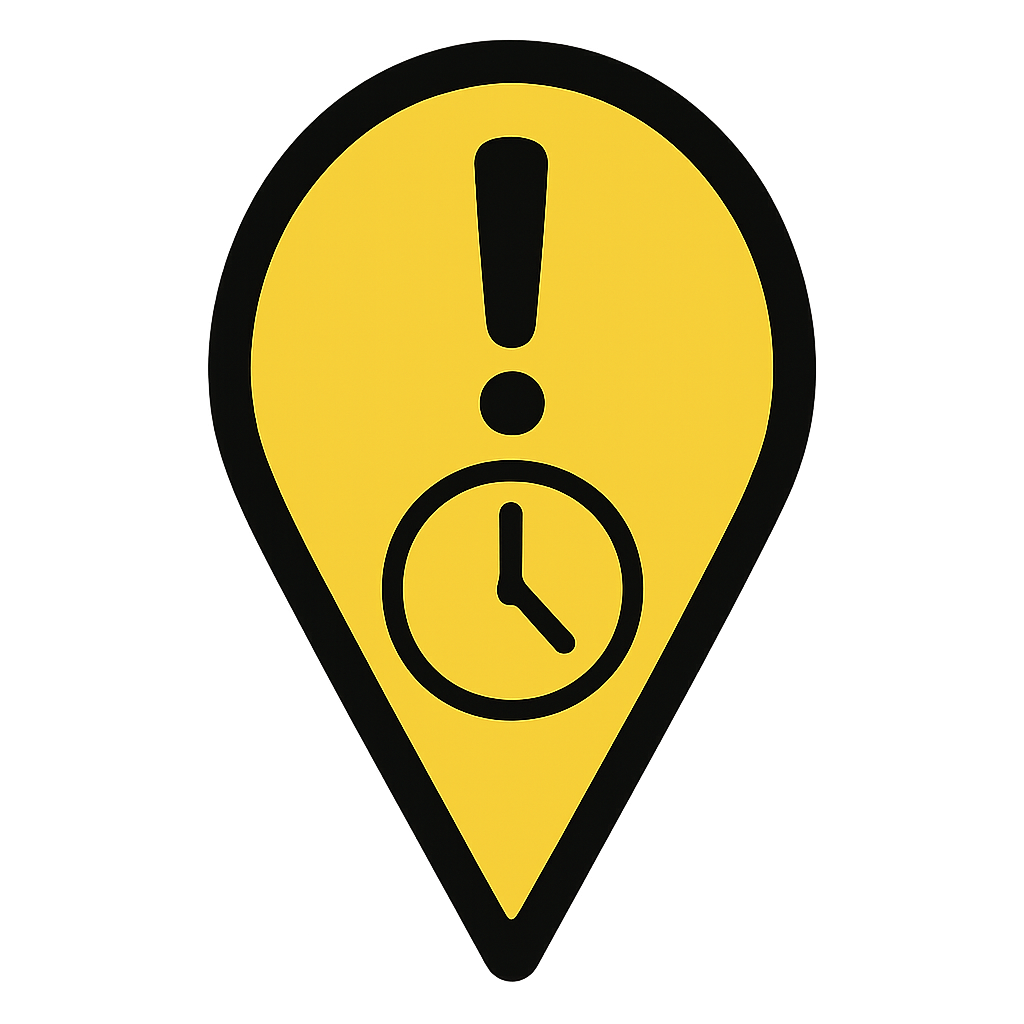} indicates prohibitive inference times (>10 mins or near-zero throughput). Baseline is the inference without any XAI method.
    MACS maintains high efficiency across context lengths.}
  \label{fig:two_side_by_side}
\end{figure*}

 As shown in Figure \ref{fig:subfigA} and Table \ref{tab:efficiency_comparison_simple}, A clear advantage of MACS is its low VRAM footprint, with only a 11\% mean increase over baseline inference; this is much more efficient than AttnLRP, which shows a 33\% increase. Attention Rollout's VRAM consumption is substantially higher, increasing by 62\% over baseline for the samples it could complete, its need to manipulate entire attention matrices leads to rapidly escalating VRAM costs and Out-of-Memory errors, especially with longer inputs. AtMan imposes negligible impact on VRAM usage (only a 1\% increase); however, this comes at the cost of greatly reduced throughput.

Regarding inference speed (Figure \ref{fig:subfigB}, Table \ref{tab:efficiency_comparison_simple}), MACS has the least impact on inference time, showing only a 23\% mean decrease in throughput. Its performance degrades gently with increasing context length and is nearly identical to the baseline for inputs under 500 tokens. This is noticeably better than AttnLRP, which sees a 53\% decrease in throughput. Attention Rollout slows inference considerably (a 59\% throughput decrease) due to its full matrix aggregations.  AtMan, being post hoc and requiring $1 +{length}_{input}$ perturbations, becomes prohibitively slow (a 78\% decrease) for long contexts, rendering it unsuitable for real-time applications.

\begin{table}[htbp]
\centering
\small

\setlength{\tabcolsep}{6pt} % Standard column separation
\renewcommand{\arraystretch}{1.2} % Standard row height
\begin{tabular}{@{}lcc@{}}
\toprule
\textbf{Method} & \textbf{Mean Peak VRAM} & \textbf{Mean Throughput} \\
                & \textbf{(MB) ↓}          & \textbf{(Tokens/Sec) ↑}    \\
\midrule
\textbf{Pure Inference} & 16088 &  7.46 \\
\midrule
Rollout           & 42649 (+62\%) & 3.04 (-59\%) \\
AtMan             & \textbf{16253 (+1\%)} & 1.66 (-78\%) \\
AttnLRP           & 24112 (+33\%) & 3.49 (- 53\%) \\
MACS (Ours)       &17998 (+11\%) & \textbf{5.69 (-23\%)} \\
\bottomrule
\end{tabular}
\caption{Efficiency comparison. For AtMan and Rollout, reported mean values are based on the subset of samples completed before encountering errors (e.g., Out-of-Memory or excessive processing time); means for other methods are over all 70 samples.↑ represents increase compared to the pure inference baseline}
\label{tab:efficiency_comparison_simple}
\end{table}

\section{Ablation Study}
To validate the design of MACS and understand the contribution of its distinct components, we conduct a series of ablation studies. We systematically remove or modify key elements of the MACS algorithm and evaluate the impact on performance. All ablations are performed on the SQuAD 2.0 subset using AUC-PR as the primary metric, as described in Section 4.1.1. The full MACS method (detailed in Section 3.2.2) serves as the baseline for comparison.
\subsection{Ablated Variants}
We evaluate the following variations of MACS:
\begin{enumerate}

    \item \textbf{w/o Redistribution}: The redistribution step  is removed. The calculation uses only the direct attention to inputs $\mathbf{a}_{I}^{(n-1,l,h)}$ instead of $\mathbf{a}_{R}^{(n-1,l,h)}$. (Section \ref{step1} Step 1)
    \item \textbf{Pooling}: Max-pooling across heads is replaced with mean-pooling/min-pooling across heads. (Section \ref{step2} Step 2)
    %\item \textbf{Mean-Pooling (vs. Max)}: Max-pooling across heads is replaced with mean-pooling. This assesses whether selecting the peak attention signal is more beneficial than averaging head contributions. (Section \ref{step2} Step 2) 
    %\item \textbf{Min-Pooling (vs. Max)}: Max-pooling is replaced with min-pooling. This explores the impact of focusing on the weakest (yet non-zero) link instead of the strongest.
    %\item \textbf{w/o Floor vector ($\alpha=1$)}: The weighted floor vector is effectively removed. This tests the importance of the floor value in preventing premature signal loss. (Section \ref{step3} Step 3)
    %\item \textbf{Varying Floor value $\alpha$ ($\alpha=0.2, \alpha=0.5$)} : This assess sensitivity to the hyperparameter.
    \item \textbf{The Floor vector }: Evaluate performance using different values for the floor hyperparameter in the $\alpha$$(1-\alpha)\mathbf{1}^N$. (Section \ref{step3} Step 3)
\end{enumerate}
\subsubsection{Discussion}
The results in Table \ref{tab:ablation_results} validate the core design choices of MACS. 

\textbf{Max-pooling is critical}, replacing it with mean-pooling or min-pooling significantly degrades mAUC-PR (from 0.601 to 0.570 and 0.443, respectively). This strongly supports our hypothesis that identifying and propagating the strongest consistent attention link via max-pooling is key to MACS's effectiveness, aligning with the idea that salient contributions are often marked by peak attention signals rather than averaged ones.

\textbf{The weighted floor vector is also essential}, its removal drops mAUC-PR to 0.546, confirming its key role in preventing the Hadamard product from prematurely nullifying token scores that only gain relevance in deeper layers. Interestingly, while the baseline is essential, the exact value of $\alpha$ has a limited impact, implying the main benefit lies in setting a non-zero floor rather than fine-tuning the weight.

\textbf{Removing the attention redistribution step had a negligible impact} on mAUC-PR in this experimental setup, possibly due to the moderate generation lengths (max 256 tokens) where its benefits for capturing distant indirect influences may be less pronounced.

\begin{table}[htbp] % [htbp] allows LaTeX to place the table optimally
\setlength{\belowcaptionskip}{0.5pt}
\centering
\setlength{\tabcolsep}{4pt} % Reduce space between columns (default is 6pt)
\renewcommand{\arraystretch}{0.9} % Reduce vertical spacing between rows (default is 1.0)

\begin{tabular}{lcc}
\toprule
\textbf{Method Variant} & \textbf{mAUC-PR}\\
\midrule
\textbf{Full MACS (Baseline)} & \textbf{0.601}  \\
\midrule
% Ablations
w/o Redistribution          & 0.600  \\
Mean-Pooling (vs Max)       & 0.570 \\
Min-Pooling (vs Max)       & 0.443 \\
w/o floor vector ($\alpha=1$)  & 0.546  \\
Floor  Vector ($\alpha=0.2$)  & 0.599 \\
Floor  Vector ($\alpha=0.5$)  & 0.599  \\
\bottomrule
\end{tabular}
\caption{Ablation study results for MACS}
% \vspace{0.2em} % Optional: Add a smaller space below the table if needed
\label{tab:ablation_results}
\end{table}

\section{Conclusion}
We have introduced Multi-Layer Attention Consistency Score (MACS), a novel heuristic for quantifying input token importance in decoder-only models. %MACS offers a distinct and practical perspective on LLM explainability compared to traditional methods.
Designed to be \textbf{lightweight, computationally efficient and usable without model modifications}.
Our empirical evaluations reveal that MACS frequently matches more sophisticated attribution techniques. These findings also support our hypothesis that measuring the consistent strength of maximal attention links across layers offers a clear and effective way to identify salient input contributions.
Consequently, MACS serves as a practical tool for obtaining rapid insights, thus contributing an efficient way toward enhancing interpretability in LLMs. %new perspective

\section{Limitations}
While MACS demonstrates compelling advantages in efficiency and offers strong empirical performance, its unique approach warrants careful consideration. MACS quantifies input contribution by measuring the consistency of maximal attention links across layers. This provides a distinct perspective on explainability compared to methods focused on input sensitivity (e.g., gradient-based) or marginal impact (e.g., perturbation-based).

The precise interpretation of this "attention consistency" score, and how it relates to or complements the insights from other established XAI paradigms, is an important area for consideration. While our results show it effectively identifies salient inputs, understanding the full implications of this consistency measure versus, for example, total information flow, requires careful interpretation based on its specific mechanism.

\section{Future Work}
This novel perspective itself opens avenues for future research into different aspects of model reasoning and what various forms of "contribution" signify (a more detailed discussion on interpreting attention consistency is provided in Section \ref{best_step} and Appendix \ref{vqa}). Other factors, such as the reliance on the underlying model's attention quality and the information selection via max-pooling, also define the scope of the current method.

Future work should therefore involve deeper theoretical analysis of this attention consistency measure, further exploring its connections to model behavior and cognitive processes across diverse tasks and Transformer architectures. Investigating its role as both a standalone diagnostic and as a component in hybrid XAI approaches also remain a promising direction.


\bibliography{acl_latex}

\appendix

\section{Experiments details}
\label{sec:appendix a}
This appendix provides further details on the datasets, metrics, and implementation specifics for the experiments presented in the main section.

The QA experiments utilized a subset of 350 question-context-answer triples from the SQuAD 2.0 dataset, selected for instances where the answer span is present in the context.
For evaluation, the \textit{original input sequence} refers to the tokenized context part of the prompt, excluding any special tokens or instructional prompt text. The ground truth ("golden answer") consists of a list of all possible tokenized answers. If multiple correct answer spans were provided in SQuAD 2.0 for a given question, our metrics were computed for each and then averaged.

\subsection{AUC-PR Calculation}
\textbf{The Area Under the Precision-Recall Curve (AUC-PR)} is our primary metric for evaluating how well an attribution method ranks the tokens from the ground-truth answer span(s) higher than non-answer tokens within the input context.
For each generation step $k$ when predicting token $t_k$, an attribution method produces a score $s_i^{(k)}$ for every token $x_i$ in the input context (length $N_{ctx}$).
Given:
\begin{itemize}
    \item A tokenized ground-truth answer span $A = \{a_1, a_2, \dots, a_M\}$.
    \item A list of $(token\_pos\_id_i, s_i^{(k)})$ pairs for the input context tokens.
\end{itemize}

We construct two vectors for the average precision score (APS) function:
\begin{itemize}
    \item $y_{true}$: A binary vector of length $N_{ctx}$. For each input context token $x_i$ at position $j$:
    $$(y_{true})_j = \begin{cases} 1 & \text{if } x_i \in A \\ 0 & \text{otherwise} \end{cases}$$
    \item $y_{score}$: A vector of the attribution scores for the corresponding input context tokens: $(y_{score})_j = s_j^{(k)}$ (score for token $x_i$ at position $j$).
\end{itemize}

The AUC-PR for a single answer span $A$ at a specific generation step $k$ is then:

$$
\text{AUC-PR}(A, k) = \texttt{APS}(\mathbf{y}_{\text{true}}, \mathbf{y}_{\text{score}})
$$

This score reflects the ability of an attribution score to correctly rank the answer tokens highly. If $any(y_{true})$ is false, the AUC-PR is treated as 0.0 for that specific case.

As a XAI method produces attribution scores at each generation step, and SQuAD 2.0 can have multiple reference answers, the final reported AUC-PR for a given sample is calculated as follows:
\begin{itemize}
    \item For each generation step during the model's response, the AUC-PR is calculated against each tokenized ground-truth answer. The mean of these AUC-PRs is taken as the step's score.
    \begingroup
    \small
    \[
    \mathrm{MeanStepAUC}^{(k)} = \frac{1}{|A|} \sum_{A_j \in A} \mathrm{AUC\text{-}PR}(A_j, k)
    \]
    \endgroup

    \item The reported AUC-PR for the entire sample is the \textbf{maximum} MeanStepAUC observed across all generation steps for that sample:
    \begingroup
    \small
    \[
    \mathrm{SampleAUC\text{-}PR} = \max_k \left( \mathrm{MeanStepAUC}^{(k)} \right)
    \]
    \endgroup
\end{itemize}
%This \textit{"Best Step"} approach acknowledges that an attribution method might highlight the answer most clearly at a particular point during the generation of the response. The overall mAUC-PR reported in the paper is the average of these SampleAUC-PR values across all 350 samples, with a 95\% confidence interval.

\subsection{Faithfulness Score Calculation}
To assess the faithfulness of the attribution scores generated by MACS and baseline methods, we adapt the "pixel flipping" (or feature removal) paradigm, commonly used in XAI, to the text domain using attention masking. This section details the calculation of the \textbf{Most Influential First (MIF)},\textbf{ Least Influential First (LIF)}, and \textbf{Symmetric Relevance Gain (SRG)} metrics based on this approach.
\subsubsection{Perturbation Strategy: Attention Masking}
For each input sample and a given attribution method, we first obtain the attribution scores (Z-scores for MACS) for all $N_{ctx}$ tokens in the input context. These scores are then used to rank the input tokens from most to least influential.

Perturbations are performed by masking tokens in the attention mechanism. Instead of removing tokens from the input sequence (which would change token positions and require re-tokenization), we modify the attention mask during the model's generation process. If a token $x_j$ is selected for "removal" based on its attribution score\footnote{For methods like MACS, AttnLRP, and Rollout that produce scores at each generation step, the attribution score used for ranking tokens for perturbation is the overall score, typically an average across all generation steps. For post hoc methods like AtMan that produce a single attribution map for the entire generated sequence, that single map is used directly.}, the attention mask is altered such that no query position can attend to key $x_j$. This effectively makes the token invisible to the attention mechanism for subsequent generation steps.

We perform these perturbations at predefined fractions $F = \{f_0, f_1, \dots, f_M\}$ of the total number of input context tokens, where $f_0 = 0\%$ (baseline, no tokens masked) and $f_M = 20\%$ in our experiments. The specific fractions used are $F = \{0.00, 0.01, 0.05, 0.10, 0.15, 0.20\}$.
\subsubsection{Performance Curves}
For each attribution method $\phi$ and each sample:
\begin{enumerate}
    \item \textbf{MIF Ordering ($\pi^{\phi}$)}: Input tokens are sorted in descending order of their attribution scores (most influential first).
    \item \textbf{LIF Ordering ($(\pi^{\phi})^r$}: Input tokens are sorted in ascending order of their attribution scores (least influential first).
    \item \textbf{Random Ordering ($\pi^{RND}$)}: Input tokens are sorted in a random order.
\end{enumerate}

For each ordering ($\pi^{\phi}$, $(\pi^{\phi})^r$, $\pi^{RND}$) and each perturbation fraction $s \in F$, we generate a new output sequence by masking the corresponding fraction $s$ of tokens. We then measure a model performance metric $v(s, \pi)$ (e.g., logit of the original next token, perplexity of the generated sequence, ROUGE-L F1, BLEU, or Semantic Similarity compared to the original unperturbed generation).

These performance values are then normalized relative to the baseline performance at $s=f_0=0\%$ perturbation (i.e., $v(f_0, \pi)$):
$$c(s, \pi) = \frac{v(s, \pi)}{v(f_0, \pi)}$$
This results in three normalized performance curves for each sample and base metric:

\begin{itemize}
    \item \textbf{$c_{MIF}(s)$}: Normalized performance curve when removing tokens by $\pi^{\phi}$.
    \item \textbf{$c_{LIF}(s)$}: Normalized performance curve when removing tokens by $(\pi^{\phi})^r$.
    \item \textbf{$c_{RND}(s)$}: Normalized performance curve when removing tokens by $\pi^{RND}$.
\end{itemize}

\subsubsection{AUC Calculation}
The Area Under the normalized Perturbation Curve (AUC) is calculated using the trapezoidal rule, normalized by the maximum perturbation fraction $f_M = 0.20$:
\begingroup
\footnotesize
\[
\begin{aligned}
\mathrm{AUC}[\pi] 
&= \frac{1}{f_M} \int_0^{f_M} c(s, \pi)\,ds \\
&\approx \frac{1}{F[-1]} \sum_{j=1}^{M} \! \left[ 
\frac{c(F[j], \pi) + c(F[j{-}1], \pi)}{2} \cdot \Delta F_j 
\right]
\end{aligned}
\]
\endgroup

where $\Delta F_j := F[j] - F[j{-}1]$ and $F[j]$ is the $j$-th fraction in our fraction array $F$.
This yields $AUC_{MIF}[\phi]$, $AUC_{LIF}[\phi]$, and $AUC_{RND}[\phi]$ for each sample.
\subsubsection{Faithfulness Metrics}
The final faithfulness metrics reported are averages for all samples.
\begin{itemize}
    \item \textbf{mMIF Score}: The average $AUC_{MIF}[\phi]$
    \item \textbf{mLIF Score}: The average $AUC_{LIF}[\phi]$
    \item \textbf{mSRG Score}: $SRG[\phi] = AUC_{LIF}[\phi] - AUC_{MIF}[\phi]$ 
\end{itemize}
The SRG for a random baseline ($SRG_{RND}$) is computed using two independent random perturbation orderings and is expected to be close to zero.

We show the comprehensive comparison across all metrics and attribution methods in Table \ref{tab:comprehensive_results}

\section{Exploring MACS's Applicability to Multi-Modal Models (Visual Question Answering)}
\label{sec:appendix b}
The core mechanism of MACS, which measures the consistency of maximal attention links across layers, is not inherently limited to text-only decoder models. It primarily requires access to layer-wise attention weights. This suggests potential applicability to other Transformer-based architectures, including multi-modal models.
\subsubsection{Motivation and Approach}
To explore this potential, we conducted a preliminary study applying MACS to a Visual Question Answering (VQA) task. A key advantage of MACS in this context is its out-of-the-box integration capability. Unlike perturbation or gradient-based XAI techniques that often require significant adaptation or architectural modifications to handle multi-modal inputs (e.g., separate handling of text and image feature perturbations, or complex gradient paths through vision and language encoders), MACS can be applied by analyzing the attention patterns within the language processing or cross-modal attention layers of the multi-modal model without changes to the underlying network.

Given the difficulty of adapting other XAI methods for a direct, rigorous comparison in this multi-modal setup without extensive engineering, our primary goal here is to demonstrate MACS's straightforward applicability and to qualitatively observe its behavior.

\subsubsection{Experimental Setup}
Similar to the QA task, we selected 200 annotated images from the Open Images Dataset V4 \cite{Kuznetsova_2020}, spanning 10 everyday object categories\footnote{Orange, Apple, Dog, Cat, Book, Laptop, Guitar, Piano, Bus, Airplane}, with 20 samples per category. To simplify the setup and reduce model load, we only included images containing exactly one annotated object. Each image was paired with the following prompt:

\begin{quote}
\texttt{What is in the image?}
\end{quote}

The goal is to evaluate if MACS effectively highlights image regions relevant to the model's predicted answer, specifically the object in the image. We report AUR PC scores calculated over the labeled masked regions. The QVA task uses the \textbf{Qwen 2.5-VL-7B} model \cite{bai2025qwen25vltechnicalreport}.
\begin{figure}[t] % Use flexible placement
  \centering 
  \vspace{0pt}
  \includegraphics[width=0.80\linewidth]{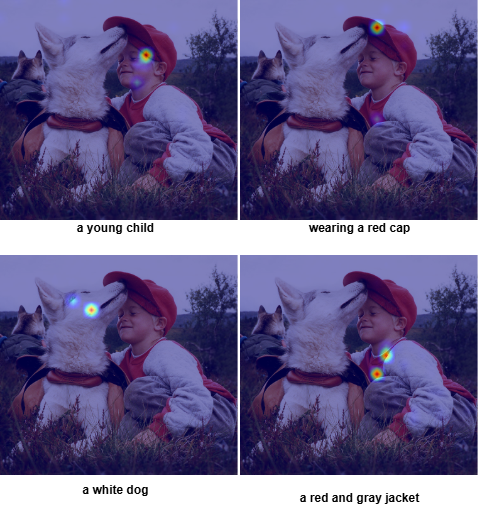}
  % You could also set the skip locally just for this figure:

  \caption{During generation, MACS dynamically highlights in the image the regions corresponding to the text as it’s being produced.
The generated text is: “The image shows a young child and a white dog, sitting together in a grassy outdoor setting. The child is wearing a red cap, a red and gray jacket, and has a backpack on.” Source: \citet{PerholsCopyright}
}
  \captionsetup{skip=1em} 
  \label{fig:enter-label}
\end{figure}

\subsubsection{Preliminary Observations}
\label{vqa}
Our application of MACS to the Qwen-VL model for this simplified VQA task yielded a mean mAUC-PR of 0.602 (averaged over 200 samples, taking the best step per sample). This score indicates that MACS is generally able to rank pixels belonging to the target object significantly higher than background pixels, demonstrating a promising level of attribution accuracy for identifying relevant visual regions.

Qualitatively, we observe a distinct pattern in MACS's attributions on these VQA samples. As illustrated in Figure \ref{fig:enter-label}.

MACS often produces sparse or "peaky" heatmaps. Instead of highlighting the entire extent of the ground-truth object, it tends to concentrate high attribution scores on a small, often well-defined, sub-region within the target object.
We hypothesize that MACS produces concentrated attribution patterns due to its core mechanism.
By measuring the consistency of maximal attention using layer-wise max-pooling and a Hadamard product, MACS highlights input features that receive consistently strong attention across layers. This results in sparse attributions focused on the most discriminative parts, rather than spreading importance evenly. 
While this "peaky" attribution leads to good ranking performance,not all ground truth object pixels receive high scores, a characteristic important for interpreting MACS outputs visually.

\begin{figure*}[t!]
    
\begin{tcbraster}[
    raster columns=1,            % We want 2 columns
    raster width=1\textwidth,   % Total width for the whole raster block
    raster equal height=rows,    % Make boxes in a row have the same height
    raster column skip=8pt,      % Space BETWEEN the columns
    raster row skip=8pt,         % Space between rows (if you have multiple)
    raster valign=top            % Align content to top within boxes
]

\begin{tcolorbox}[
  colback=gray!10,           % light gray background
  colframe=ForestGreen!50!white,
  boxrule=0.5pt,
  title={%
    \small Q: In what country is Normandy located?\\
    \small G: \underline{Normandy}
  },
]
\small
The Normans (Norman: Nourmands; French: Normands; Latin: Normanni) were the people
who in the 10th and \colorbox{red!5}{11th} centuries gave their name to Normandy,
a region in \colorbox{red!70}{France}. They were descended from Norse (“Norman”
comes from “Norseman”) raiders and pirates from Denmark, Iceland and Norway who,
under their leader Rollo, agreed to swear fealty to King Charles III of West
Francia. Through generations of assimilation and mixing with the native Frankish
and Roman-Gaulish populations, their descendants would gradually merge with the
Carolingian-based cultures of West Francia. The distinct cultural and ethnic
identity of the Normans emerged initially in the first half of the 10th century,
and it continued to evolve over the succeeding centuries.
\end{tcolorbox}

\begin{tcolorbox}[
  colback=gray!10,           % light gray background
  colframe=ForestGreen!50!white,
  boxrule=0.5pt,
  title={%
    \small Q: In what country is Normandy located?\\
    \small G: Normandy \underline{is}
  },
]
\small
The Normans (Norman: Nourmands; French: Normands; Latin: Normanni) were the people
who in the 10th and 11th centuries gave their name to \colorbox{red!70}{Normandy},
\colorbox{red!15}{a region in} \colorbox{red!35}{France}. They were descended from Norse (“Norman”
comes from “Norseman”) raiders and pirates from Denmark, Iceland and Norway who,
under their leader Rollo, agreed to swear fealty to King Charles III of West
Francia. Through generations of assimilation and mixing with the native Frankish
and Roman-Gaulish populations, their descendants would gradually merge with the
Carolingian-based cultures of West Francia. The distinct cultural and ethnic
identity of the Normans emerged initially in the first half of the 10th century,
and it continued to evolve over the succeeding centuries.
\end{tcolorbox}

\begin{tcolorbox}[
  colback=gray!10,           % light gray background
  colframe=ForestGreen!50!white,
  boxrule=0.5pt,
  title={%
    \small Q: In what country is Normandy located?\\
    \small G: Normandy is \underline{located}
  },
]
\small
The Normans (Norman: Nourmands; French: Normands; Latin: Normanni) were the people
who in the 10th and 11th centuries gave their name to Normandy\colorbox{red!5}{,}
a\colorbox{red!15}{region in} \colorbox{red!70}{France}. They were descended from Norse (“Norman”
comes from “Norseman”) raiders and pirates from Denmark, Iceland and Norway who,
under their leader Rollo, agreed to swear fealty to King Charles III of West
Francia. Through generations of assimilation and mixing with the native Frankish
and Roman-Gaulish populations, their descendants would gradually merge with the
Carolingian-based cultures of West Francia. The distinct cultural and ethnic
identity of the Normans emerged initially in the first half of the 10th century,
and it continued to evolve over the succeeding centuries.
\end{tcolorbox}

\begin{tcolorbox}[
  colback=gray!10,           % light gray background
  colframe=ForestGreen!50!white,
  boxrule=0.5pt,
  title={%
    \small Q: In what country is Normandy located?\\
    \small G: Normandy is located \underline{in}
  },
]
\small
The Normans (Norman: Nourmands; French: Normands; Latin: Normanni) were the people
who in the 10th and 11th centuries gave their name to Normandy,
a region \colorbox{red!10}{in} \colorbox{red!70}{France}. They were descended from Norse (“Norman”
comes from “Norseman”) raiders and pirates from Denmark, Iceland and Norway who,
under their leader Rollo, agreed to swear fealty to King Charles III of West
Francia. Through generations of assimilation and mixing with the native Frankish
and Roman-Gaulish populations, their descendants would gradually merge with the
Carolingian-based cultures of West Francia. The distinct cultural and ethnic
identity of the Normans emerged initially in the first half of the 10th century,
and it continued to evolve over the succeeding centuries.
\end{tcolorbox}

\end{tcbraster}

\caption{MACS demonstrating anticipatory attention on a QA example. The heatmap shows MACS attribution scores on the input context prior to the model generating the answer "France". High consistency scores (darker red) on "France" in the context indicate MACS identifies the answer span before its generation by the model (Q: Question, G: Generated tokens).}
\label{fig:gshowcase}
\end{figure*}

\begin{table*}[!t]
\makeatletter
\setlength{\@fptop}{0pt}
\resizebox{\textwidth}{!}{%
\begin{tabular}{l@{\hspace{6pt}}l@{\hspace{6pt}}r@{\hspace{6pt}}r@{\hspace{6pt}}r@{\hspace{6pt}}r@{\hspace{6pt}}r}
\toprule
\textbf{Output Type} & \textbf{Metric} & \textbf{MACS} & \textbf{AttnLRP} & \textbf{Rollout} & \textbf{AtMan} & \textbf{Random} \\
\midrule
\multirow{1}{*}{Main Task} 
& AUC-PR & \textbf{$0.601\pm0.033$} & $0.565\pm0.030$ & $0.147\pm0.021$ & $0.315\pm0.032$ & $0.113\pm0.012$ \\
\midrule

\multirow{3}{*}{Mean Logits} 
& mMIF (AUC) & $0.919\pm0.006$ & $0.915\pm0.006$ & $0.967\pm0.005$ & $0.945\pm0.006$ & $-$ \\
& mLIF (AUC) & $1.003\pm0.002$ & $1.005\pm0.002$ & $0.995\pm0.003$ & $0.964\pm0.005$ & $-$ \\
& mSRG       & $0.084\pm0.006$ & $0.090\pm0.006$ & $0.028\pm0.005$ & $0.019\pm0.006$ & $-0.003\pm0.003$ \\
\midrule

\multirow{3}{*}{Perplexity} 
& mMIF (AUC) & $1.114\pm0.013$ & $1.120\pm0.014$ & $1.040\pm0.010$ & $1.046\pm0.010$ & $-$ \\
& mLIF (AUC) & $0.996\pm0.004$ & $0.994\pm0.004$ & $1.002\pm0.005$ & $1.025\pm0.008$ & $-$ \\
& mSRG       & $-0.118\pm0.013$ & $-0.126\pm0.013$ & $-0.039\pm0.009$ & $-0.021\pm0.010$ & $0.003\pm0.005$ \\
\midrule

\multirow{3}{*}{ROUGE-L F1} 
& mMIF (AUC) & $0.619\pm0.021$ & $0.609\pm0.021$ & $0.823\pm0.019$ & $0.783\pm0.019$ & $-$ \\
& mLIF (AUC) & $0.933\pm0.013$ & $0.932\pm0.013$ & $0.906\pm0.014$ & $0.838\pm0.018$ & $-$ \\
& mSRG       & $0.315\pm0.021$ & $0.323\pm0.022$ & $0.082\pm0.019$ & $0.055\pm0.020$ & $-0.006\pm0.011$ \\
\midrule

\multirow{3}{*}{BLEU} 
& mMIF (AUC) & $0.571\pm0.020$ & $0.558\pm0.019$ & $0.780\pm0.022$ & $0.718\pm0.021$ & $-$ \\
& mLIF (AUC) & $0.903\pm0.017$ & $0.902\pm0.016$ & $0.869\pm0.019$ & $0.784\pm0.021$ & $-$ \\
& mSRG       & $0.332\pm0.021$ & $0.344\pm0.021$ & $0.089\pm0.021$ & $0.066\pm0.022$ & $-0.011\pm0.012$ \\
\midrule

\multirow{3}{*}{Semantic Sim.} 
& mMIF (AUC) & $0.758\pm0.020$ & $0.752\pm0.020$ & $0.892\pm0.014$ & $0.870\pm0.015$ & $-$ \\
& mLIF (AUC) & $0.962\pm0.009$ & $0.962\pm0.009$ & $0.944\pm0.011$ & $0.906\pm0.013$ & $-$ \\
& mSRG       & $0.205\pm0.019$ & $0.210\pm0.019$ & $0.051\pm0.014$ & $0.036\pm0.015$ & $-0.004\pm0.009$ \\
\bottomrule
\end{tabular}
}%
\caption{Comprehensive Comparison of Attribution Methods. All values are averaged across all samples.}
\label{tab:comprehensive_results}
\end{table*}

\end{document}